\pdfoutput=1
\documentclass[11pt]{article}

\usepackage[preprint]{acl}

\usepackage{times}
\usepackage{latexsym}

\usepackage[T1]{fontenc}
\usepackage[utf8]{inputenc}

\usepackage{microtype}

\usepackage{inconsolata}

\usepackage{graphicx}

\title{Examining Language Modeling Assumptions Using an Annotated Literary Dialect Corpus}

\author{Craig Messner \\
  Center for Digital Humanities \\
  Johns Hopkins University \\
  \texttt{cmessne4@jhu.edu} \\\And
  Tom Lippincott \\
  Center for Digital Humanities \\
  Johns Hopkins University \\
  \texttt{tom.lippincott@jhu.edu} \\}

\begin{document}
\maketitle
\begin{abstract}
We present a dataset of 19th century American literary orthovariant tokens with a novel layer of human-annotated dialect group tags designed to serve as the basis for computational experiments exploring literarily meaningful orthographic variation.  We perform an initial broad set of experiments over this dataset using both token (BERT) and character (CANINE)-level contextual language models. We find indications that the "dialect effect" produced by intentional orthographic variation employs multiple linguistic channels, and that these channels are able to be surfaced to varied degrees given particular language modelling assumptions. Specifically, we find evidence showing that choice of tokenization scheme meaningfully impact the type of orthographic information a model is able to surface.
\end{abstract}

\section{Introduction}

Orthographic variation, the deviation from one system of spelling in favor of another, occurs due to a range of intentional and unintentional motivations. Unintentional variation may occur when a writer misspells a word relative to their intended system, or when an optical character recognition system misidentifies a particular character. Intentional deviations are instead used to create a desired political or literary effect \cite{sebba2007spelling}. For example, adhering to a system of simplified spelling may signal one's dedication to egalitarian politics, while embedding a literary character's speech in a particular orthographic form may signal an authorial desire to present that character as belong to a particular race, class, region or gender \cite{ives1971theory} \cite{jones1999strange}.

This latter class of intentional variations proves especially diverse. Supported by the availability of surrounding context and reader-familiar stereotypes of speech, literary orthographic edits are frequently unsystematic ("eye dialect") or not fully beholden to phonetics or morphology \cite{krapp1925english}. Instead, the means by which they convey a "dialect effect" is likely multidimensional.

We present a dataset that includes a novel human-annotated layer of dialect family tags designed to support investigations into these varied signalling pathways. We perform an initial set of experiments and discover indications that literary orthographic variation communicates its dialect effect by modifying information along multiple axes: word-level semantics, context-level semantics, and character edits. In the spirit of previous work investigating the phonetic \cite{agirrezabal-etal-2023-hidden}, semantic \cite{rahman-etal-2023-token} and contextual \cite{ethayarajh-2019-contextual} information token and character level models capture, we also provide analysis of the literary orthographic understanding of these model types. We additionally offer evidence that character-level models distinguish between intentional literary orthovariants and constructed unintentional variants.
%We present a series of experiments over a dataset of intentionally orthovariant tokens drawn from the literature of 19th century America. These experiments use a battery of character, token and type level language models to demonstrate how literary orthographic variation communicates a "dialect effect" by modifying information along multiple axes: word-level semantics, context-level semantics, and character edits. 

%A reader encountering a literary character uttering "pahentage" might use their contextual %knowledge of the surrounding utterance, as well as the author's intimation that this personage is white and from the American south, to decode the token to "parentage". They may then further expect this character to use the orthovariant "suh" in place of "sir". However, should the author instead choose to use a different variant form (say, "sa") the reader's contextual knowledge still allows for decoding this token as a "southern form" \cite{walpole1974eye}.

%In the course of doing so we provide:
%\begin{itemize}
%    \item A novel layer of "Dtag" dialect group annotations 
%    \item Analysis of the literary orthographic capability of token and character level models, in the spirit of previous work investigating the types of phonetic \cite{agirrezabal-etal-2023-hidden}, semantic \cite{rahman-etal-2023-token} and contextual \cite{ethayarajh-2019-contextual} information each captures.
%    \item Evidence that character-level contextual embedding models distinguish between intentional literary variants and constructed unintentional variants.
%\end{itemize}

\section{Experiments}
\subsection{Setup}
\textbf{Data.} The data for the following experiments consists of 4032 orthovariant tokens paired with their standard forms and sentence-level context, drawn from a 19th century American literature subset of the Project Gutenberg corpus. This corpus is further described in \cite{messner-lippincott-2024-pairing}. Messner extended the tag set by providing an additional "Dtag" drawn from a set of 31 possibilities, indicating the dialect form ascribed to each observed token.

%We tokenize the texts, split them into sentences, and identify plausibly orthovariant tokens by discounting those included in WordNet or the Brown corpus. We also disqualified those with capitalization or numeric characters. Author 1 identified the actually orthovariant tokens in this subset, and assigned them a "standard" modern English token

Messner used the authorially intended subject-position of speaking characters to assign Dtags to tokens. As a result, the Dtag set mostly represents perceived race, nationality, and region. The most populous category (1726 tokens) is the backwoods (BW) tag which combines samples from white-identified northeastern, western and central plains characters. These subcategories are of BW are often only subtly disjoint; distinguishing them is likely to cause confusion. Other frequent tags include AA (African American: 653), AR (intentionally archaic: 549), GA (Gaelic: 336) and DE (German: 220).

\textbf{Models.} We employ six models for the following experiments. One, fastText-pretrained \cite{mikolov-etal-2018-advances} is a subword-aware type level embedding model provided by Facebook and trained on CommonCrawl. We use four pretrained token-level contextual models. Two, BERT-large-uncased and BERT-base-uncased \cite{devlin-etal-2019-bert} use WordPiece tokenization, while CANINE-c and CANINE-s \cite{clark-etal-2022-canine} are character-level, with the latter utilizing an additional subword loss function during training. Finally, BERT-forced is BERT-base-uncased configured to encode input strings using only single character WordPiece tokens.

\subsection{Procedure}

\textbf{Embeddings: the absolute set.} We truncate the dataset, keeping only samples that fit the BERT-forced limit of 512 characters, resulting in 3871 observed-standard pairs. For each pair we generate four additional synthetic tokens:
\begin{enumerate}
    \item \textbf{rev:} The standard word in reversed character order. Ex: circus \verb|->| sucric 
    \item \textbf{ocr:} A mutated version of the standard word produced using the nlpaug \cite{ma2019nlpaug} library's OCR error engine. Ex: circus \verb|->| cikcos
    \item \textbf{swp:} The standard word with a single character swap Ex: circus \verb|->| icrcus
    \item \textbf{rnd:} The standard word with a randomly mutated single character Ex: circus \verb|->| circun
\end{enumerate}
We collect embeddings for this full token set. For the type-level model, we embed each individual word. For the contextual models, we insert each variant into the context sentence in turn, embed the full sentence, and extract the set of embeddings that represent the target word. For the BERT family of models, we use the last four hidden layers of the model as the embedding values, while for CANINE we use the final hidden layer. If the target word is embedded as more than one subword or character we mean pool the sub-embeddings to generate a final word embedding.

\textbf{Data augmentation: the relative set.} We use these embeddings to produce additional datapoints consisting of the difference between the embedding of a token's standard form and the embedding of each of the variant forms. Similar to the analogy test of \cite{mikolov-etal-2013-linguistic}, we use these relative datapoints to investigate a given model's ability to preserve the intuition that similar types of orthographic transformation should produce similar differences in $n$-dimensional space.

%hat similar types of orthographic transformation, relative to their standard forms, should produce similar forms of difference in $n$-dimensional space. %Ideally, models will preserve the analogical intuition that the difference between a standard token and constructed variant is different from the difference between a standard and its corresponding observed token. Similarly, an ideal model would preserve the intuition that the difference between two distinct tokens that share a Dtag and their standard equivalents is more similar than the difference between another set of tokens and variants with a distinct Dtag.

For each model, we cluster the relative and absolute sets using $k$-means clustering for each $k \in \{1, ..,20\}$. \footnote{Code and data for these experiments can be found at \url{https://github.com/comp-int-hum/orthography-embedding-clustering}}

\subsection{Evaluation}
We use the following measures to evaluate the efficacy of a given $k$ clustering. 

\textbf{Purity.} We calculate purity \cite{manning2008introduction} over the clustering of the full relative token set to gain insight into each model's ability to distinguish between synthetic and observed variants. We also calculate it over the absolute and relative sets of only the observed token to track how well the models cluster embeddings or embedding differences that bear the same Dtag.

%\textbf{Purity} quantifies a given $k$ clustering's ability to group each token type into distinct clusters:
%\begin{equation}
%    p = \frac{\sum_{0}^{k} max(|O_k|,|S_k|,|R_k|,|C_k|,|N_k|,|W_k|)}{totaltokens}
%\end{equation}

%Where $O$ indicates the $k$ cluster's associated observed nonstandard tokens, $S$ standard tokens, $R$ rev tokens, $C$ ocr tokens, $N$ rnd tokens, and $W$ swp tokens.  We also apply purity to examine the efficacy of Dtag clustering.

\textbf{Overall accuracy and SO accuracy} evaluate a given $k$ clustering's ability to group token variants from the same datapoint into the same cluster. Overall accuracy is the average percentage of correct groupings of all elements of a datapoint into a single cluster. SO accuracy is the average percentage of correct groupings of only the standard and observed tokens into a single cluster.

\textbf{Cluster semantic coherency} measures the overall semantic similarity of the tokens gathered into a cluster $k$. We calculate this using the average pointwise cosine similarity of the Word2Vec \cite{mikolov2013efficient} embeddings of each token in a cluster. To support this we train a Word2Vec model on the full corpus using the Gensim \cite{rehurek_lrec} library.

\textbf{Cluster Mphone similarity} measures the phonetic similarity of the tokens gathered into a cluster $k$. We calculate this using the average pointwise Levenshtein Distance (LD) of the Metaphone \cite{Philips1990HangingOT} encoded version of each token in the cluster. A lower score indicates that the members of the cluster are more phonetically similar.

%We calculate this using the average pointwise similarity score for all tokens in a cluster using a given similarity function, in this case the cosine similarity of Word2Vec \cite{mikolov2013efficient} embeddings for each token pair. To support this, we train a Word2Vec model on the full corpus using the Gensim \cite{rehurek_lrec} library.

\begin{figure}[h!]
    \centering
    \includegraphics[width=0.8\linewidth]{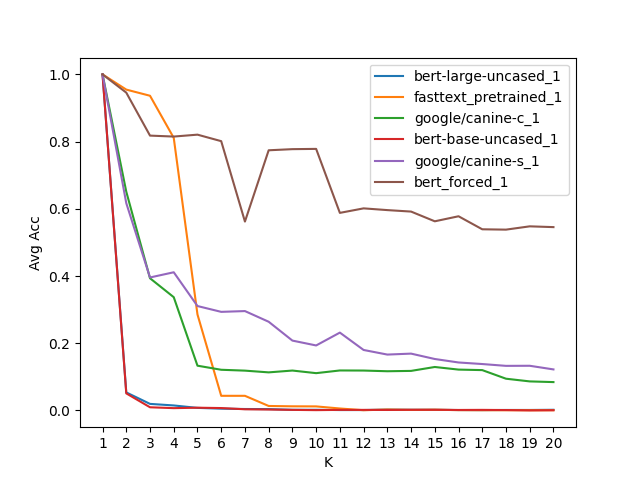} \vspace{.1cm}
    \includegraphics[width=0.8\linewidth]{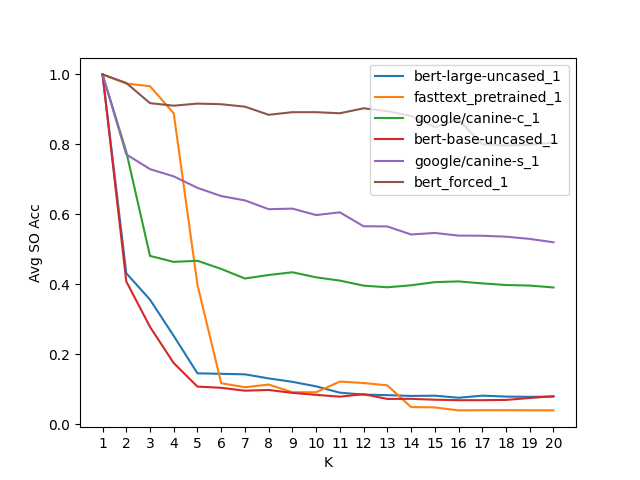}
    \caption{Full absolute set (T), SO absolute set (B) accuracy  by $k$.}
    \label{fig:ftacc}
\end{figure}

\section{Results and Discussion}

\subsection{Evaluating absolute}

\textbf{Only BERT-forced consistently embeds all variants into a similar region.} Figure~\ref{fig:ftacc} demonstrates that all of the models except for BERT-forced perform uniformly poorly on both overall accuracy across all $k$, barring the uninformative case where $k < 6$. 

\textbf{The models that perform best on SO accuracy are character-level.}

Again barring the uninformative $K<6$ cases, BERT-forced, CANINE-s and CANINE-c best separate observed-standard pairings from other tokens in their datapoints (Figure~\ref{fig:ftacc}). Analysis of their shared error reveals that both models perform poorly on a set of high Levenshtein Distance (LD) edit pairs (average LD 2.67). Correspondingly, their shared correct token transformation set has a lower average LD of 1.66. BERT-forced performs better on higher LD transformations, with average correct and error set average LD of 2.2 and 1.9 respectively, implying that BERT-forced preserves difference information beyond character edits.

\begin{figure}[h!]
    \centering
    \includegraphics[width=0.8\linewidth]{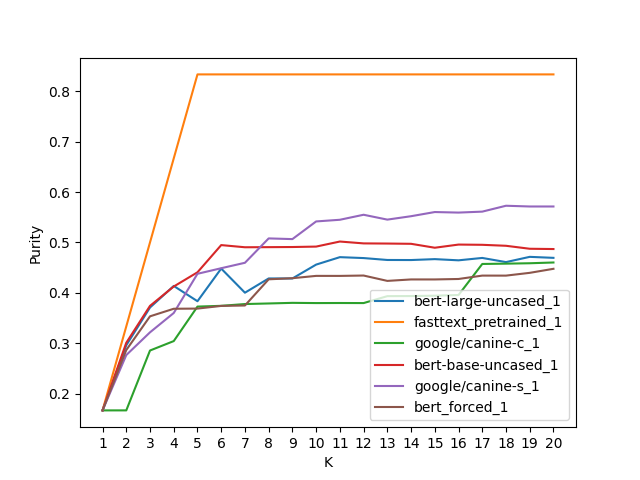} \vspace{.1cm}
    \includegraphics[width=0.8\linewidth]{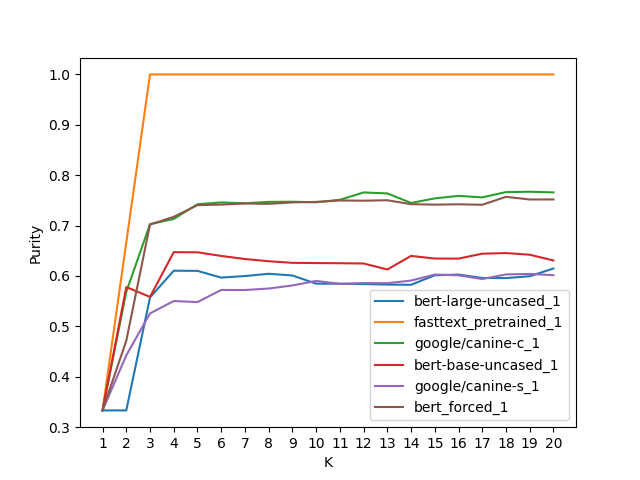}
    \caption{Purity across the full relative set (T) and across non order-swapped tokens (B)}
    \label{fig:ftpurity}
\end{figure}

\subsection{Evaluating relative}

\textbf{Of the character and token level models, the CANINE series most distinctly separates constructed and non-constructed variants into clusters.}  The type-level fastText-pretrained model most accurately separates the variants (Figure~\ref{fig:ftpurity}). However closer examination reveals that it does not separate individual constructed forms, instead grouping them into a single cluster.
%\begin{figure*}[t!]
%    \centering
%    \includegraphics[width=0.48\linewidth]{images/combined/Dtag_embed_purity.png} \hfill
%    \includegraphics[width=0.48\linewidth]{images/combined/Dtag_diff_purity.png}
%    \caption{Dtag purity over the obv token embedding (L) dtag purity over the std-obv token embedding (R)}
%    \label{fig:dtpurity}
%\end{figure*}
Notably, character-level models treat order-swapped tokens as functionally similar, while token-level models do not. Removing the rev and swp tokens benefits all models, but overall benefits character-level models the most. Ultimately, this indicates that character-level models preserve information about the distinctions between standard/constructed token differences and standard/observed token differences. It also implies that they rely to a greater degree on the character-edit information stream of the dialect effect to make this determination.

\subsection{Evaluation in the light of Dtag and semantic information} \label{dtag}

\begin{figure}[h!]
    \centering
    \includegraphics[width=0.8\linewidth]{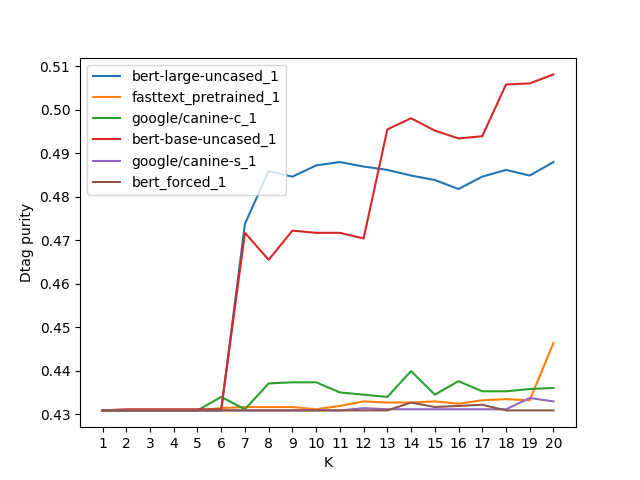} \vspace{.1cm}
    \includegraphics[width=0.8\linewidth]{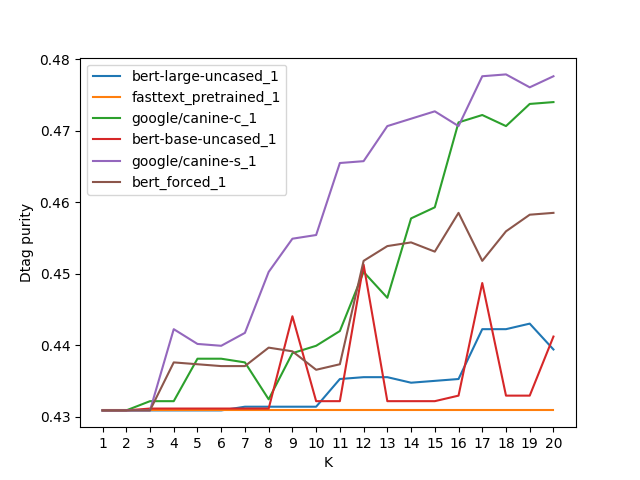}
    \caption{Dtag purity over the obv token embedding (T) Dtag purity over the std-obv relative set (B)}
    \label{fig:dtpurity}
\end{figure}

\textbf{High performance on Dtag clustering relies on a mixture of word-semantic, context-semantic and character edit information.} As $K$ increases, BERT-base performs best on absolute and CANINE-s on relative (Figure~\ref{fig:dtpurity}). However, both models ultimately only reach purity scores of $\sim .5$, in part at least due to the dominance of the BW tag. Investigating the proportion of individual dtags on a per-cluster basis at the jointly performant $k$=17 reveals how both models capture a partial mix of these signals. 

\textbf{Certain cluster compositions have potential literary significance.} 
Clusters 3 and 7 of the CANINE-s relative set contain high proportions of both AA (African-American) and WS (White Southern) labeled tokens (Table~\ref{csdtag}).

\begin{table}[h!]
\centering
\tabcolsep=0.13cm
\begin{tabular}{llrrrc}
\hline
\textbf{K} & \textbf{Count}  & \textbf{aa} & \textbf{bw} & \textbf{ws} &\textbf{Mphone} \\ \hline
3          & 160             & 0.36        & 0.19        & 0.24      &3.1  \\ \hline
7          & 53            & 0.59        & 0.08        & 0.19      &2.5   \\ \hline
\end{tabular}
\caption{Excerpted Dtag proportions and Mphone similarities of CANINE-s relative set clusters at $K=17$}
\label{csdtag}
\vspace{-.3cm}
\end{table}
Both clusters have low word-level semantic coherence scores (.25 and .27 respectively), consistent with the bulk of of the other clusters at $k=17$, indicating that this grouping likely does not emerge from word-semantics. This Dtag clustering is particularly striking, as it suggests that period authors took a position on the debate surrounding the origins of southern speech \cite{bonfiglio2010race}. 

\begin{table}[h]
\begin{tabular}{l|ll}
\hline
\textbf{Edit} & \textbf{Standard} & \textbf{Observed}\\
\hline
er \verb|->| ah  & after   & aftah   \\
er \verb|->| a   & rather  & ratha  \\
r \verb|->| '      & quarters & qua'ters   \\
\hline
\end{tabular}
\caption{Characteristic edits and examples shared by AA and WS in CANINE-s relative clusters 3 and 7}
\label{table:SamplesAAWS}
\vspace{-.3cm}
\end{table}

The edits shared by WS and AA in these clusters (Table~\ref{table:SamplesAAWS}) largely impact "r"-related graphemes, demonstrating that this clustering likely occurs due to character edit information. Notably, cluster 3 ranks as less sel-similar than cluster 7 by average Mphone LD. Upon inspection, cluster 3 contains a wider variety of "r"-related edits, including r \verb|->| y and r \verb|->| w. In combination with the somewhat more broad distribution of Dtags found in cluster 3, this implies that these "r" edits are somehow nearer to the sorts of "r" edits characteristic of other Dtag groupings found in the cluster, potentially for contextual reasons. 

A similar type of distribution also occurs over GA (Gaelic) tokens. Clusters 8 and 12 contain uniquely high proportions of GA tokens (.27 each) while retaining typically low word-level semantic coherence (.24 and .25). Inspection of the tokens reveals that these clusters collect a variety of edits to the "i" and "e" graphemes. However, unlike the WS and AA clusters examined above, both share similar Mphone LD averages of 3.1 and 3.3 respectively. This may signal that these "i" and "e" transformations are more broadly indicative of a variety of dialect contexts.

\textbf{Context semantics in part determines accurate literary variant clustering.} Notably, the BERT-base absolute set at $k$=17 centralizes clusters around different tags while diffusing WS and GA tokens. For example, cluster 14 has a significantly higher proportion (.33) of DE (German) tagged tokens than any DE-containing cluster found in the CANINE-s relative set.

\begin{table}[h]
\begin{tabular}{l|ll}
\hline
\textbf{Edits} & \textbf{Standard} & \textbf{Observed}\\
\hline
b \verb|->| p  & poem   & boem   \\
-g   & blooming  & bloomin  \\
u \verb|->| oo      & hunters & hoonters   \\
f \verb|->| v      & falls & valls   \\
\hline
\end{tabular}
\caption{Characteristic edits and examples of DE tagged tokens in BERT-base absolute cluster 14}
\label{table:SamplesDE}
\vspace{-.3cm}
\end{table}
The DE tokens in this cluster (Table~\ref{table:SamplesDE}) represent a diverse set of edits, including one (-g in the terminal position) associated with numerous Dtags, including BW (Backwoods) and AA. Given this cluster's low semantic coherence (.34), a likely conclusion is that this cluster emerges due to the similarity of orthographic contexts in which these tokens appear -- say an utterance laden with other characteristic DE edits.

\textbf{Low performance on Dtag clustering correlates with high word-semantic cluster coherence in the relative set.} For example, BERT-large relative contains multiple clusters with semantic coherency $>.5$, while CANINE-s relative has only one cluster with a score $>.4$. This implies these models favor preserving word-semantic analogical relationships over character edit and context semantics relationships, destabilizing the blend of information needed to successfully cluster over Dtags.

\section{Conclusions and Further Work}
These experiments offer indications that the dialect effect presented by literary orthographic variation utilizes multiple channels of information: contextual semantics, word semantics and character edits. They also offer evidence that while both contextual token and character level language models can capture all of these aspects, they do so unevenly, justifying further work on the best combination of their information streams.

\section{Limitations}
The primary limitation of this study emerges from the data. Beyond the inherent limitation of self-restriction to works by 19th century American authors, the coherence of a given observed token and its assigned Dtag is also limited by the inventory of tags chosen. Authors of this period grant their characters multidimensional subject-positions that are reasonably described by but not fully reducible to the granularity of tags like WS and AA. Analysis done in a Dtag-to-cluster direction where the assigned tags are taken as full ground truth limits access to these subtleties. 

\bibliography{custom, anthology}

\begin{thebibliography}{18}
\providecommand{\natexlab}[1]{#1}

\bibitem[{Agirrezabal et~al.(2023)Agirrezabal, Boldsen, and Hollenstein}]{agirrezabal-etal-2023-hidden}
Manex Agirrezabal, Sidsel Boldsen, and Nora Hollenstein. 2023.
\newblock \href {https://doi.org/10.18653/v1/2023.cawl-1.2} {The hidden folk: Linguistic properties encoded in multilingual contextual character representations}.
\newblock In \emph{Proceedings of the Workshop on Computation and Written Language (CAWL 2023)}, pages 6--13, Toronto, Canada. Association for Computational Linguistics.

\bibitem[{Bonfiglio(2010)}]{bonfiglio2010race}
Thomas~Paul Bonfiglio. 2010.
\newblock \emph{Race and the rise of standard American}, volume~7.
\newblock Walter de Gruyter.

\bibitem[{Clark et~al.(2022)Clark, Garrette, Turc, and Wieting}]{clark-etal-2022-canine}
Jonathan~H. Clark, Dan Garrette, Iulia Turc, and John Wieting. 2022.
\newblock \href {https://doi.org/10.1162/tacl_a_00448} {Canine: Pre-training an efficient tokenization-free encoder for language representation}.
\newblock \emph{Transactions of the Association for Computational Linguistics}, 10:73--91.

\bibitem[{Devlin et~al.(2019)Devlin, Chang, Lee, and Toutanova}]{devlin-etal-2019-bert}
Jacob Devlin, Ming-Wei Chang, Kenton Lee, and Kristina Toutanova. 2019.
\newblock \href {https://doi.org/10.18653/v1/N19-1423} {{BERT}: Pre-training of deep bidirectional transformers for language understanding}.
\newblock In \emph{Proceedings of the 2019 Conference of the North {A}merican Chapter of the Association for Computational Linguistics: Human Language Technologies, Volume 1 (Long and Short Papers)}, pages 4171--4186, Minneapolis, Minnesota. Association for Computational Linguistics.

\bibitem[{Ethayarajh(2019)}]{ethayarajh-2019-contextual}
Kawin Ethayarajh. 2019.
\newblock \href {https://doi.org/10.18653/v1/D19-1006} {How contextual are contextualized word representations? {C}omparing the geometry of {BERT}, {ELM}o, and {GPT}-2 embeddings}.
\newblock In \emph{Proceedings of the 2019 Conference on Empirical Methods in Natural Language Processing and the 9th International Joint Conference on Natural Language Processing (EMNLP-IJCNLP)}, pages 55--65, Hong Kong, China. Association for Computational Linguistics.

\bibitem[{Ives(1971)}]{ives1971theory}
Sumner Ives. 1971.
\newblock A theory of literary dialect.
\newblock \emph{A various language: Perspectives on American dialects}, pages 145--177.

\bibitem[{Jones(1999)}]{jones1999strange}
Gavin Jones. 1999.
\newblock \emph{Strange talk: The politics of dialect literature in Gilded Age America}.
\newblock Univ of California Press.

\bibitem[{Krapp(1925)}]{krapp1925english}
George~Philip Krapp. 1925.
\newblock \emph{The English Language in America}, volume~1.
\newblock Century Company, for the Modern language association of America.

\bibitem[{Ma(2019)}]{ma2019nlpaug}
Edward Ma. 2019.
\newblock Nlp augmentation.
\newblock https://github.com/makcedward/nlpaug.

\bibitem[{Manning(2008)}]{manning2008introduction}
Christopher~D Manning. 2008.
\newblock Introduction to information retrieval.

\bibitem[{Messner and Lippincott(2024)}]{messner-lippincott-2024-pairing}
Craig Messner and Thomas Lippincott. 2024.
\newblock \href {https://aclanthology.org/2024.latechclfl-1.26} {Pairing orthographically variant literary words to standard equivalents using neural edit distance models}.
\newblock In \emph{Proceedings of the 8th Joint SIGHUM Workshop on Computational Linguistics for Cultural Heritage, Social Sciences, Humanities and Literature (LaTeCH-CLfL 2024)}, pages 264--269, St. Julians, Malta. Association for Computational Linguistics.

\bibitem[{Mikolov(2013)}]{mikolov2013efficient}
Tomas Mikolov. 2013.
\newblock Efficient estimation of word representations in vector space.
\newblock \emph{arXiv preprint arXiv:1301.3781}.

\bibitem[{Mikolov et~al.(2018)Mikolov, Grave, Bojanowski, Puhrsch, and Joulin}]{mikolov-etal-2018-advances}
Tomas Mikolov, Edouard Grave, Piotr Bojanowski, Christian Puhrsch, and Armand Joulin. 2018.
\newblock \href {https://aclanthology.org/L18-1008} {Advances in pre-training distributed word representations}.
\newblock In \emph{Proceedings of the Eleventh International Conference on Language Resources and Evaluation ({LREC} 2018)}, Miyazaki, Japan. European Language Resources Association (ELRA).

\bibitem[{Mikolov et~al.(2013)Mikolov, Yih, and Zweig}]{mikolov-etal-2013-linguistic}
Tomas Mikolov, Wen-tau Yih, and Geoffrey Zweig. 2013.
\newblock \href {https://aclanthology.org/N13-1090} {Linguistic regularities in continuous space word representations}.
\newblock In \emph{Proceedings of the 2013 Conference of the North {A}merican Chapter of the Association for Computational Linguistics: Human Language Technologies}, pages 746--751, Atlanta, Georgia. Association for Computational Linguistics.

\bibitem[{Philips(1990)}]{Philips1990HangingOT}
Lawrence Philips. 1990.
\newblock \href {https://api.semanticscholar.org/CorpusID:59912108} {Hanging on the metaphone}.

\bibitem[{Rahman et~al.(2023)Rahman, Sakib, Faisal, and Anastasopoulos}]{rahman-etal-2023-token}
Md~Mushfiqur Rahman, Fardin~Ahsan Sakib, Fahim Faisal, and Antonios Anastasopoulos. 2023.
\newblock \href {https://doi.org/10.18653/v1/2023.mrl-1.6} {To token or not to token: A comparative study of text representations for cross-lingual transfer}.
\newblock In \emph{Proceedings of the 3rd Workshop on Multi-lingual Representation Learning (MRL)}, pages 67--84, Singapore. Association for Computational Linguistics.

\bibitem[{{\v R}eh{\r u}{\v r}ek and Sojka(2010)}]{rehurek_lrec}
Radim {\v R}eh{\r u}{\v r}ek and Petr Sojka. 2010.
\newblock {Software Framework for Topic Modelling with Large Corpora}.
\newblock In \emph{{Proceedings of the LREC 2010 Workshop on New Challenges for NLP Frameworks}}, pages 45--50, Valletta, Malta. ELRA.
\newblock \url{http://is.muni.cz/publication/884893/en}.

\bibitem[{Sebba(2007)}]{sebba2007spelling}
Mark Sebba. 2007.
\newblock \emph{Spelling and society: The culture and politics of orthography around the world}.
\newblock Cambridge University Press.

\end{thebibliography}
%\bibliography{custom}

\end{document}